\documentclass[10pt, a4paper]{article}
\usepackage{lrec}
\usepackage{multibib}
\newcites{languageresource}{Language Resources}
\usepackage{graphicx}
\usepackage{colortbl}
\usepackage{tabularx}
\usepackage{bm}
\usepackage{multirow}
\usepackage{soul}

\usepackage{epstopdf}
\usepackage[latin1]{inputenc}

\usepackage{hyperref}
\usepackage{xstring}

\title{Performance Impact Caused by Hidden Bias of Training Data\\ for Recognizing Textual Entailment}

\name{Masatoshi Tsuchiya}

\address{Toyohashi University of Technology \\
         1--1 Hibarigaoka, Tempaku-cho, Toyohashi, Aichi, Japan \\
         tsuchiya@imc.tut.ac.jp}

\abstract{%
  The quality of training data is one of the crucial problems when a
  learning-centered approach is employed.
  This paper proposes a new method to investigate the quality of a large
  corpus designed for the recognizing textual entailment (RTE) task.
  The proposed method, which is inspired by a statistical hypothesis
  test, consists of two phases: the first phase is to introduce the
  predictability of textual entailment labels as a null hypothesis which
  is extremely unacceptable if a target corpus has no hidden bias, and
  the second phase is to test the null hypothesis using a Naive Bayes
  model.
  The experimental result of the Stanford Natural Language Inference
  (SNLI) corpus does not reject the null hypothesis.  Therefore, it
  indicates that the SNLI corpus has a hidden bias which allows
  prediction of textual entailment labels from hypothesis sentences even
  if no context information is given by a premise sentence.
  This paper also presents the performance impact of NN models for RTE
  caused by this hidden bias.
  \\ \newline }

\begin{document}

\maketitleabstract

\section{Introduction}
\label{sec:intro}

The quality of a training data is one of the crucial problems when a
learning-centered approach including neural network (NN) is employed.
\cite{reidsma_CL2008} demonstrated that annotation errors of the dialog
act corpus, which follows a certain systematic pattern, mislead the
learning result of Bayesian network.
\cite{zhang_ICLR2017} described that the capacity of a NN model is large
enough for brute-force memorizing the entire data set, even if its
labels are random.  Thus, an influence for a NN model caused by a
certain systematic pattern may become more serious than the influence
for other learning-centered methods.

Both a method to improve the quality of a training data and a metric to
evaluate its reliability are important.
As the former method, majority vote of human annotators was widely used
\cite{sabou_LREC2014,bowman_EMNLP2015,alkhatib_COLING2016}
As the latter metric, many kind of inter-annotator agreement metrics
based on multiple human annotation results were employed, because direct
assessment of data quality is difficult
\cite{craggs_CL2005,artstein_CL2008,mathet_COLING2012}.

This paper proposes a new empirical method to investigate a quality of a
large corpus designed for the recognizing textual entailment (RTE) task
\cite{condoravdi_NAACL2003,bos_HLT2005,maccartney_IWCS2009,semeval2014}.
The proposed method, which is inspired by a statistical hypothesis test,
assesses a quality of a target corpus directly and does not depend on
multiple human annotation results unlike the existing metrics.
The proposed method consists of two phases: the first phase is to
introduce the predictability of textual entailment (TE) labels as a null
hypothesis which is extremely unacceptable when a target corpus has no
hidden bias, and the second phase is to test the null hypothesis for the
target corpus using a Naive Bayes (NB) model.


In order to demonstrate the efficiency of the proposed method, we
investigate two RTE corpora: the Stanford Natural Language Inference
(SNLI) corpus \cite{bowman_EMNLP2015} and the Sentences Involving
Compositional Knowledge (SICK) corpus \cite{marelli_LREC2014}.
Although the experimental result of the SICK corpus rejects the null
hypothesis, the result of the SNLI corpus does not reject it.
Thus, the SNLI corpus has a hidden bias to allow prediction of TE labels
from hypothesis sentences even if no context information is given by a
premise sentence.
The other experiment shows that this hidden bias causes the risk that a
NN model for RTE works as a entirely different model than its
constructor expects.

The major contributions of this paper are following three points:
\begin{itemize}
  \itemsep=0pt
  \item This paper proposes a new empirical method to reveal a hidden
	bias of a large RTE corpus (Section~\ref{sec:predictability}).
  \item This paper applies the proposed method on the SNLI corpus and the
	SICK corpus, and reveals the hidden bias of the SNLI corpus (Section~\ref{sec:experiment}).
  \item This paper also presents that this hidden bias causes the risk that
	a NN model proposed for RTE works as a entirely different model
	than its constructor expects (Section~\ref{sec:discussion}).
\end{itemize}

\begin{figure}
  \centering
  \small
  \begin{tabular}{|lp{0.85\columnwidth}|}
    \hline
    $s_{1}$ & Two boys are swimming in the pool. \\
    $s_{2}$ & Two girls are playing basketball. \\
    $s_{3}$ & Two women are swimming in the pool. \\
    \hline
    $s_{h}$ & Two children are swimming in the pool. \\
    \hline
  \end{tabular}
  \caption{Example sentences of RTE. The textual entailment label of
  $s_h$ is determinable if and only if context information is given by a
  premise sentence.}
  \label{fig:RTE_example}
\end{figure}

\begin{figure*}
  \centering
  \begin{minipage}[t]{0.97\columnwidth}
    {\centering
    \small
    \begin{tabular}{|l|r|r|r|}
      \hline
      Predicted & \multicolumn{3}{c|}{Corpus labels} \\ \cline{2-4}
      labels & Entailment & Neutral & Contradiction \\
      \hline
      Entailment & \multicolumn{1}{>{\columncolor[gray]{0.8}}r|}{2275} & 644 & 706 \\ \hline
      Neutral & 508 & \multicolumn{1}{>{\columncolor[gray]{0.8}}r|}{1976} & 563 \\ \hline
      Contradiction & 585 & 599 & \multicolumn{1}{>{\columncolor[gray]{0.8}}r|}{1968} \\
      \hline
    \end{tabular}}
    \\ \normalsize
    (a) The confusion matrix obtained by the TE label prediction model
    trained and tested on the SNLI corpus.  It tries to predict an
    appropriate TE label for each individual hypothesis sentence.
  \end{minipage}
  \hfil
  \begin{minipage}[t]{0.97\columnwidth}
    {\centering
    \small
    \begin{tabular}{|l|r|r|r|}
      \hline
      Predicted & \multicolumn{3}{c|}{Corpus labels} \\ \cline{2-4}
      labels & Entailment & Neutral & Contradiction \\
      \hline
      Entailment & 3 & 3 & 2 \\ \hline
      Neutral & \multicolumn{1}{>{\columncolor[gray]{0.8}}r|}{1411} & \multicolumn{1}{>{\columncolor[gray]{0.8}}r|}{2790} & \multicolumn{1}{>{\columncolor[gray]{0.8}}r|}{718} \\ \hline
      Contradiction & 0 & 0 & 0 \\
      \hline
    \end{tabular}}
    \normalsize
    (b) The confusion matrix obtained by the TE label prediction model
    trained and tested on the SICK corpus.  It simply tries the major TE
    label `neutral' for almost all hypothesis sentences without
    prediction.
  \end{minipage}
  \caption{Confusion matrices of TE Label Prediction Models}
  \label{fig:confusion_matrices}
\end{figure*}

\section{Proposed Method}
\label{sec:predictability}

This section describes the detail of the proposed method, which consists
of two phases.
%

\subsection{Predictability of TE Labels without Premise Sentences}
\label{subsec:method}

The first phase of the proposed method is to derive a null hypothesis
which is extremely unacceptable when a target corpus has no hidden bias.
We focus the task definition of the target corpus for this phase.

\cite{semeval2014} defined RTE as a task to partition relationships
between a premise sentence and a hypothesis sentence into three
categories: entailment, neutral and contradiction.
Consider the example sentences shown in Figure~\ref{fig:RTE_example}.
When $s_1$ is given as a premise sentence and $s_h$ is given as a
hypothesis sentence, the relationship between $s_1$ and $s_h$ is
labeled entailment.
The relationship between $s_2$ and $s_h$ is labeled neutral, and
the relationship between $s_3$ and $s_h$ is labeled contradiction.
These examples indicate that the TE label is determinable if and only if
context information is given by a premise sentence.
Based on this observation, the null hypothesis of the proposed method is
defined as follows:
\begin{description}
  \item[Definition of the null hypothesis] \ \\
	     The TE label of the hypothesis sentence is determinable without the
	     premise sentence.
\end{description}
Because the null hypothesis looks extremely unacceptable, it is
appropriate to reveal a hidden bias of the target RTE corpus.

\subsection{TE Label Prediction Model}
\label{subsec:model}

The second phase of the proposed method is to test statistical
significance of the null hypothesis for a target corpus.
This phase requires two models: the first model is the statistical model
of the null hypothesis, henceforth referred to as the {\itshape TE label
prediction model}, and the second model is the statistical model of the
alternative hypothesis, henceforth referred to as the baseline model.

The TE label prediction model is a model which predicts TE labels for
hypothesis sentences without context information of premise sentences.
This paper employs a multinomial NB model \cite{wang_ACL2012} as the TE
label prediction model, which is defined by the following equation:
\begin{equation}
  \hat{y} = \mathop{\mathrm{argmax}}_{y}P(y)\prod_{i=1}^{n}P(x_i|y),
\end{equation}
where $y$ is a TE label and $x_i$ is a feature.
This paper simply employs all word unigrams of hypothesis sentences as
features.

The baseline model assigns TE labels for hypothesis sentences when no
information is given by either premise sentences or hypothesis sentences
but only the TE label distribution $P(y)$ of the target corpus is
available.
In such case, it is reasonable to assign the TE label which occurs most
frequently in the target corpus for all hypothesis sentences.  This
baseline assignment is defined as follows:
\begin{equation}
  \breve{y} = \mathop{\mathrm{argmax}}_{y}P(y)
\end{equation}

%
If there is a statistically significant difference between the TE label
prediction model and the baseline model, the null hypothesis is not
rejected for the target corpus, and it indicates that the target corpus
contains a hidden bias.
Otherwise, the null hypothesis is rejected for the target corpus,

\begin{table}
  \centering
  \small
  \begin{tabular}{l|r|r}
    \hline
    Corpus & \multicolumn{1}{c|}{TE label prediction model} & \multicolumn{1}{c}{Baseline model} \\
    \hline
    SNLI & {\bfseries 63.3\%} & 34.3\% \\
    SICK & 56.7\% & 56.7\% \\
    \hline
  \end{tabular}
  \caption{Performance of the TE label prediction model trained and
  tested on two RTE corpora.  The TE label prediction model trained and
  tested on the SNLI corpus achieves statistically significant accuracy
  than the baseline model.}
  \label{tbl:label_prediction_results}
\end{table}

\section{Experiment}
\label{sec:experiment}

This section presents the detailed experimental conditions and the
experimental results.
The highlight of these results is that the TE label prediction model
achieves 63.3\% accuracy for the SNLI corpus.


Table~\ref{tbl:label_prediction_results} shows the performances of the
TE label prediction models trained and tested on two RTE
corpora\footnote{\cite{scikit-learn} is employed to implement the TE
label prediction model.}.
The TE label prediction model, which is trained on the SNLI training
hypothesis sentences and their TE labels, achieves 63.3\% accuracy on
the SNLI test hypothesis sentences without premise sentences.
The baseline model based on the SNLI TE label distribution achieves
34.3\% accuracy on the same hypothesis sentences.
The sign test indicates that there is a statistically significant
difference between these models ($p=5.7e^{-202}$).
On the other hand, the performance of the TE label prediction model
trained and tested on the SICK corpus is close to the performance of the
baseline model (56.7\%).  The sign test indicates that there is no
statistically significant difference between these models ($p=0.65$).

Figure~\ref{fig:confusion_matrices} clearly shows the difference between
the behavior of the TE label prediction model trained on the SNLI corpus
and the model trained on the SICK corpus.
The left matrix is obtained by the model trained and tested on the SNLI
corpus, and the right matrix is obtained by the model trained and tested
on the SICK corpus.
The model of the SICK corpus simply tries the major TE label `neutral'
for almost all hypothesis sentences without prediction, although the
model of the SNLI corpus tries to predict an appropriate TE label for
each individual hypothesis sentence.

These results indicate that the null hypothesis is rejected for the SICK
corpus, but it is not rejected for the SNLI corpus.
Therefore, hypothesis sentences of the SNLI corpus have a hidden bias to
allow prediction of their TE labels without premise sentences.

\begin{table}
  \centering
  \small
  \begin{tabular}{l|r@{ }r|r@{ }r}
    \hline
    & \multicolumn{2}{c|}{$E_{e}$} & \multicolumn{2}{c}{$H_{e}$} \\
    \hline
    Entailment    & 2,275 & (36.6\%) & 1,093 & (30.3\%) \\
    Neutral       & 1,976 & (31.8\%) & 1,243 & (34.5\%) \\
    Contradiction & 1,968 & (31.6\%) & 1,269 & (35.2\%) \\
    \hline
    Total & 6,219 & (63.3\%) & 3,605 & (36.7\%) \\
    \hline
  \end{tabular}
  \caption{Empirical classification of the SNLI corpus using the TE label prediction model}
  \label{tbl:SNLI_easy_hard_classification}
\end{table}

\begin{table*}
  \centering
  \small
  \begin{tabular}{l|r@{ }l|r|r|r}
    \hline
    Model & \multicolumn{2}{c|}{Related models}
	& \multicolumn{1}{c|}{$E_{e}\cup H_{e}$}
        & \multicolumn{1}{c|}{$E_{e}$}
	& \multicolumn{1}{c}{$H_{e}$} \\
    \hline
    Parallel LSTM model & 76.3\% & {\footnotesize \cite{mou_EMNLP2016}} & 76.8\% & 87.8\% & 57.8\% \\
    Sequential LSTM model & 80.9\% & {\footnotesize \cite{rocktaschel_ICLR2016}} & 81.4\% & 90.1\% & 65.6\% \\
    \hline
  \end{tabular}
  \caption{Performance of the NN models for RTE.
  Performance drops from the empirical easy test set $E_e$
  to the empirical hard test set $H_e$ are observed for both NN
  models.}
  \label{tbl:RTE_performance}
\end{table*}

\begin{table*}
  \centering
  \small
  \begin{tabular}{l|r|r|r}
    \hline
    Model & \multicolumn{1}{c|}{$E_{e}\cup H_{e}$}
        & \multicolumn{1}{c|}{$E_{e}$}
	& \multicolumn{1}{c}{$H_{e}$} \\
    \hline
    Parallel LSTM model & 54.1\% & 66.0\% & 33.7\% \\
    Sequential LSTM model & 48.6\% & 56.7\% & 34.7\% \\
    \hline
  \end{tabular}
  \caption{Performance of the NN models for RTE, when all words of
  premise sentences are replaced by unknown word symbols.  Although all
  context information is removed by this replacement, both NN models
  achieve obviously higher performance than the chance ratio for the
  empirical easy test set $E_e$.  This result indicates that both NN
  models do not work as RTE models for $E_e$, but work as TE label
  prediction models for $E_e$.}
  \label{tbl:RTE_random_premise_results}
\end{table*}

\begin{table*}[!tb]
  \centering
  \footnotesize
  \begin{tabular}{l||r@{ }r|r@{ }r|r@{ }r||r@{ }r|r@{ }r|r@{ }r}
    \hline
    & \multicolumn{6}{c||}{SNLI} & \multicolumn{6}{c}{SICK} \\ \cline{2-13}
    & \multicolumn{2}{c|}{Training} & \multicolumn{2}{c|}{Development} & \multicolumn{2}{c||}{Test}
    & \multicolumn{2}{c|}{Training} & \multicolumn{2}{c|}{Development} & \multicolumn{2}{c}{Test} \\
    \hline
    Entailment & 183,416 & (33.4\%) & 3,329 & (33.8\%) & 3,368 & (34.3\%) & 1,299 & (28.9\%) & 144 & (28.8\%) & 1,414 & (28.7\%) \\
    Neutral & 182,764 & (33.3\%) & 3,235 & (32.9\%) & 3,219 & (32.8\%) & 2,536 & (56.4\%) & 282 & (56.4\%) & 2,793 & (56.7\%) \\
    Contradiction & 183,187 & (33.4\%) & 3,278 & (33.3\%) & 3,237 & (33.0\%) & 665 & (14.8\%) &  74 & (14.8\%) &   720 & (14.6\%) \\
    \hline
    Total & 549,367 & & 9,842 & & 9,824 & & 4,500 & & 500 & & 4,927 & \\
    \hline
  \end{tabular}
  \caption{Statistics of TE labels of two RTE corpora.  Although TE labels of the
  SNLI corpus are balanced, TE labels of the SICK corpus are not balanced.}
  \label{tbl:RTE_corpora_statistics}
\end{table*}

\section{Discussion}
\label{sec:discussion}

As described in Section~\ref{sec:experiment}, hypothesis sentences of
the SNLI corpus have a hidden bias to allow prediction of their TE
labels without premise sentences.
The magnitude of the performance impact caused by the hidden bias is
important, because the SNLI corpus is widely used as training data by
many NN models for RTE
\cite{bowman_EMNLP2015,rocktaschel_ICLR2016,yin_TACL2016,mou_ACL2016,wang_NAACL2016,liu_ACL2016,liu_EMNLP2016,cheng_EMNLP2016,parikh_EMNLP2016,sha_COLING2016}.
This section discusses the performance impact of the NN models caused by
the hidden bias.

\subsection{Empirical Classification of SNLI Corpus}
\label{subsec:empirical_division}

The test pairs of the SNLI corpus are classified into two subsets using
the TE label prediction model trained on the SNLI corpus.
The first subset is the empirical easy test set $E_{e}$, which consists
of all test pairs whose TE labels are predicted correctly by the TE
label prediction model.
The second subset is the empirical hard test set $H_{e}$, which consists
of the rest pairs.
Table~\ref{tbl:SNLI_easy_hard_classification} shows the classification
result.
63.3\% test pairs of the SNLI corpus were classified as $E_{e}$, and the
rest pairs were classified as $H_{e}$.

\subsection{Definitions of NN Models for RTE}

Two NN models are prepared to evaluate performance impacts caused by the
hidden bias.
The first model (henceforth denoted as the parallel LSTM model) was
proposed by \cite{bowman_EMNLP2015} for RTE, and was evaluated on the
performance difference between RTE corpora by \cite{mou_EMNLP2016}.
This model is defined by the following equations.
\begin{eqnarray*}
  \bm{h}_{p,i} &=& \mbox{LSTM}_{p}(W_{e}x_{p,i} + W_{hp}\bm{h}_{p,i-1}) \\
  \bm{h}_{h,i} &=& \mbox{LSTM}_{h}(W_{e}x_{h,i} + W_{hh}\bm{h}_{h,i-1}) \\
  \bm{l}_{1} &=& \tanh(W_{1}[\bm{h}_{p,|\bm{x_{p}}|}, \bm{h}_{h,|\bm{x_{h}}|}] + B_{1}) \\
  \bm{l}_{2} &=& \tanh(W_{2}\bm{l}_{1} + B_{2}) \\
  \bm{l}_{3} &=& \tanh(W_{3}\bm{l}_{3} + B_{3}) \\
  \bm{y} &=& \mathrm{softmax}(\bm{l}_{3})
\end{eqnarray*}
The first step is to convert a premise sentence $\bm{x}_{p}$ and a
hypothesis sentence $\bm{x}_{h}$ into embedding vectors using the word
embedding matrix $W_{e}$, which is initialized with the 300-dimension
reference GloVe vectors \cite{pennington_EMNLP2014}.
The second step is to convert embedding vectors into two 100-dimension
sentence vectors with LSTMs, and they are concatenated into a
200-dimension vector.
The remaining steps are to predict a TE label with three tanh fully connected
layers and then to apply the softmax function.

The second model (henceforth denoted as the sequential LSTM model),
which was proposed by \cite{rocktaschel_ICLR2016} for RTE, is defined as
follows.
\begin{eqnarray*}
  \bm{h}_{p,i} &=& \mbox{LSTM}_{p}(W_{e}x_{p,i} + W_{hp}\bm{h}_{p,i-1}) \\
  \bm{h}_{h,0} &=& \mbox{LSTM}_{h}(W_{hh}\bm{h}_{p,|\bm{x_{p}}|}) \\
  \bm{h}_{h,i} &=& \mbox{LSTM}_{h}(W_{e}x_{h,i} + W_{hh}\bm{h}_{h,i-1}) \\
  \bm{l} &=& \tanh(W_{l}\bm{h}_{h,|\bm{x_{h}}|} + B_{l}) \\
  \bm{y} &=& \mathrm{softmax}(\bm{l})
\end{eqnarray*}
In the second model, two LSTMs are sequentially connected. Thus, it is
possible to consider that the memory cells of these LSTMs are directly
modeling a recognition process unlike the parallel LSTM model.
%
%
All vectors of the sequential LSTM model are 100-dimension.
Although \cite{rocktaschel_ICLR2016} proposed the variants with
attentions between a premise sentence and a hypothesis sentence, the
attention-less model is employed in this experiment, because of its
simplicity.

\subsection{Performance Impact of NN Models for RTE}

This subsection presents the big performance drop of the NN models
caused by the hidden bias.

Table~\ref{tbl:RTE_performance} shows the experimental results of these
NN models trained and tested on the SNLI corpus.
Although both NN models achieve high accuracy for the whole test set and
for the empirical easy test set $E_{e}$, they achieve drastic low
accuracy for the empirical hard test set $H_{e}$.
These performance drops mean that a large portion of the high accuracy
achieved by both NN models benefits from the empirical easy test set
$E_{e}$.

Table~\ref{tbl:RTE_random_premise_results} shows the performance which
is achieved by the same NN models when all words of premise sentences
are replaced by unknown word symbols.
Because this replacement removes all context information from premise
sentences, thus the performance of the NN models must drop close to the
chance ratio, if the NN models decide TE labels based on context
information of premise sentences.
Despite this expectation, both NN models achieve obviously higher
performance than the chance ratio for $E_e$ (36.8\%, shown in
Table~\ref{tbl:SNLI_easy_hard_classification}).
This result indicates that both NN models do not work as RTE models
for $E_e$, but work as TE label prediction models for $E_e$.
This behavior of NN models for $E_e$ must be entirely different than
their constructor expected.

\subsection{Comparison of SNLI and SICK corpora}

The SNLI and SICK corpora, are entirely similar in their sentence
domains, English scene descriptions.
Both of them use the Flickr30k corpus \cite{young_TACL2016} as origins
of their sentences.
It is also exhibited by the small differences of sentence token mean
counts as shown in Table~\ref{tbl:snli_sick}.
The second is about their vocabulary.  The out-of-vocabulary (OOV)
ratio of SICK test pairs is 0.15\%, when words of SNLI training pairs
are regarded as known.  This small OOV ratio indicates that SICK test
pairs and SNLI training pairs are quite close from the view point of
their vocabulary.

\begin{table}
  \centering
  \begin{tabular}{l|r|r}
    \hline
    & \multicolumn{1}{c|}{SNLI} & \multicolumn{1}{c}{SICK} \\
    \hline
    \# of training pairs & 55k & 4,500 \\
    \# of development pairs & 10k & 500 \\
    \# of test pairs & 10k & 4,927 \\
    Premise mean token count
	& 14.1 
	& 9.8 
	\\
    Hypothesis mean token count
	& 8.3 
	& 9.5 
	\\ 
    Vocabulary size of training pairs & 36,427 & 2,178 \\
    Vocabulary size of test pairs & 6,548 & 2,188 \\
    OOV ratio of test pairs
	& 0.25\% 
	& 0.32\% 
	\\
    \hline
  \end{tabular}
  \caption{Comparison of the SNLI and SICK corpora.  Both corpora are
  extremely similar from the view point of descriptive statistics.}
  \label{tbl:snli_sick}
\end{table}

The SNLI and SICK corpora are different in the method of composing
sentences.  Hypothesis sentences of the SNLI corpus are composed by
human workers, but all sentences of the SICK corpus are derived from
original sentences using hand-crafted rules.
We think that this difference may be a cause of the hidden bias revealed
by this paper.
%

\section{Conclusion}
\label{sec:conclusion}

This paper proposes a new empirical method to investigate the quality of
large RTE corpus.
The proposed method consists of two phases: the first phase is to
introduce the predictability of TE labels as a null hypothesis, and the
second phase is to test the null hypothesis using a NB model.
The proposed method reveals a hidden bias of the SNLI corpus, which
allows prediction of TE labels from hypothesis sentences without context
information given by premise sentences.

This paper also presents that this hidden bias makes a large performance
impact on the NN models for RTE.
The experimental result shows that a large portion of the high accuracy
achieved by the NN models benefits from the hidden bias.
The other experimental result shows that a NN model trained on the SNLI
corpus does not work as an RTE model, but works as a TE label prediction
model, when biased test pairs are given.
These results arise a risk that a complex NN model works as an entirely
different model than its constructor expects.

%
%
%
%
%

\section{Acknowledgments}
A part of this research was supported by JSPS KAKENHI Grant No.~15K12097.
I would like to express my sincere appreciation to
Dr. Mitsuo Yoshida and Dr. Adam Meyers for their valuable comments.

\section{Bibliographical References}
\bibliographystyle{lrec}
\bibliography{main}

\end{document}